\pdfoutput=1  
\documentclass[11pt]{article}

\usepackage[preprint]{acl}

\usepackage{times}
\usepackage{latexsym}

\usepackage[T1]{fontenc}

\usepackage[utf8]{inputenc}

\usepackage{microtype}

\usepackage{inconsolata}

\usepackage{graphicx}

\usepackage[utf8]{inputenc} 
\usepackage[T1]{fontenc}    
\usepackage{hyperref}       
\usepackage{url}            
\usepackage{booktabs}       
\usepackage{amsfonts}       
\usepackage{nicefrac}       
\usepackage{microtype}      
\usepackage{xcolor}         
\usepackage{makecell}
\usepackage{multirow}

\usepackage{amsmath}
\usepackage{amssymb}
\usepackage{mathtools}
\usepackage{amsthm}
\usepackage{diagbox}
\usepackage{bbm}
\usepackage{pifont}

\usepackage{bm}
\usepackage{graphicx}
\usepackage{multirow}
\usepackage{colortbl}
\usepackage{rotating}
\usepackage{makecell}
\usepackage{wrapfig}
\usepackage{enumitem}
\usepackage{pifont}
\usepackage{makecell}

\usepackage[ruled,vlined,linesnumbered]{algorithm2e}

\usepackage[most]{tcolorbox}

\usepackage{multicol}
\usepackage{setspace}

\definecolor{best}{rgb}{1,  0.851,  0.4}
\definecolor{second}{rgb}{0.557,  0.663,  0.859}

\definecolor{groupA}{RGB}{240, 245, 255} 
\definecolor{groupB}{RGB}{245, 240, 255} 
\definecolor{headergray}{RGB}{230, 230, 230} 

\definecolor{groupC}{RGB}{248, 248, 248} 
\definecolor{groupD}{RGB}{240, 240, 240} 
\definecolor{bestRed}{RGB}{255, 0, 0}    
\definecolor{bestOrange}{RGB}{255, 140, 0} 

\usepackage{hyperref}
\SetCommentSty{footnotesize}
\SetKwComment{Comment}{$\triangleright$ }{}
\SetSideCommentRight

\title{Rationale-Grounded In-Context Learning for Time Series Reasoning \\ with Multimodal Large Language Models}

\author{
  Qingxiang Liu\textsuperscript{1}, 
  Zhiqing Cui\textsuperscript{1},
  Xiaoliang Luo\textsuperscript{2},
  Yuqian Wu\textsuperscript{1},
  Zhuoyang Jiang\textsuperscript{1},
  \\
  \textbf{Huaiyu Wan\textsuperscript{3},} 
  \textbf{Sheng Sun\textsuperscript{4},}
  \textbf{Lvchun Wang\textsuperscript{2},}
  \textbf{Wei Yu\textsuperscript{2},}
  \textbf{Yuxuan Liang\textsuperscript{1,}\thanks{Corresponding Authors.}}
  \\
  \textsuperscript{1}The Hong Kong University of Science and Technology (Guangzhou) \\
  \textsuperscript{2}China Mobile (Jiangxi) Virtual Reality Technology Co., Ltd.\\
  \textsuperscript{3}School of Computer and Information Technology, Beijing Jiaotong University\\
  \textsuperscript{4} Institute of Computing Technology, Chinese Academy of Sciences\\
  \texttt{\{qingxiangliu737@gmail.com, yuxliang@outlook.com\}}
}

\begin{document}
\maketitle
\begin{abstract}
The underperformance of existing multimodal large language models for time series reasoning lies in the absence of rationale priors that connect temporal observations to their downstream outcomes, which leads models to rely on superficial pattern matching rather than principled reasoning.
We therefore propose the rationale-grounded in-context learning for time series reasoning, where rationales work as guiding reasoning units rather than post-hoc explanations, and develop the \texttt{RationaleTS} method.
Specifically, we firstly induce label-conditioned rationales, composed of reasoning paths from observable evidence to the potential outcomes.
Then, we design the hybrid retrieval by balancing temporal patterns and semantic contexts to retrieve correlated rationale priors for the final in-context inference on new samples.
We conduct extensive experiments to demonstrate the effectiveness and efficiency of our proposed \texttt{RationaleTS} on three-domain time series reasoning tasks.
We will release our code for reproduction.


\end{abstract}

\section{Introduction}

Time series reasoning is fundamental to decision making in ubiquitous real-world domains, such as air pollution warning~\cite{cui2025augur}, transportation management~\cite{yu2024rethinking}, and healthcare monitoring~\cite{liu2023large}.
The reasoning performance hinges on not only extrapolating historical trends, but also modeling the interaction among multiple variables over time and analyzing how temporal contexts correspond to future outcomes~\cite{jiang2025timexl}. 
Thus, time series reasoning actually covers diverse tasks of prediction, classification, and anomaly detection, where the generated results must be supported by the evidence from the historical horizons and temporal contexts~\cite{kong2025position,ni2025harnessing}.

\begin{figure*}[!t]
    \centering
    \includegraphics[width=0.82\linewidth]{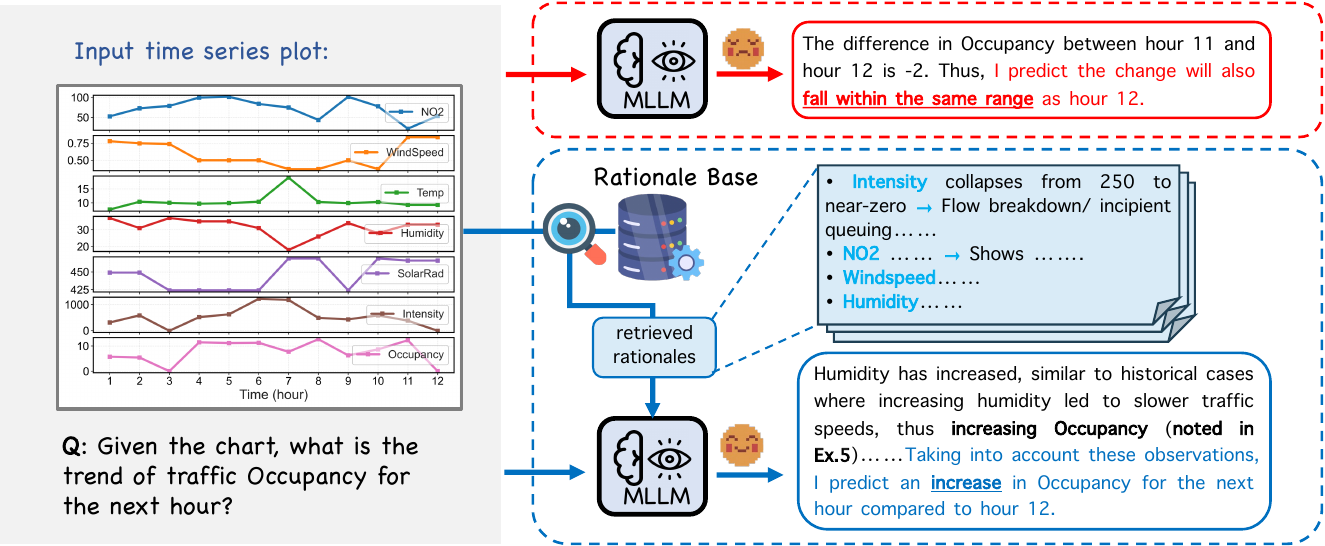}
    \caption{Comparison in time series reasoning paradigms with MLLMs (\textcolor{red}{red part}) and rationale-grounded in-context learning in \texttt{RationaleTS} (\textcolor{blue}{blue part}). In MLLMs the prediction outcome is generated by pattern extrapolation, while in \texttt{RationaleTS}, rationales provide reasoning priors connecting observations and implications, for the in-context learning on new samples.}
    \label{fig_intro}
\end{figure*}

Recent advances in Multimodal Large Language Models (MLLMs) have motivated their use for time series reasoning by converting numerical sequences into visual charts, promising jointly perceiving temporal patterns and generating natural language explanations in a unified framework~\cite{liu2025picture,zhong2025time,wang2025chattime}.
However, despite the improved modality alignment compared with converting numerical data into textual tokens in LLMs~\cite{jin2023time,liu2024time,chang2025llm4ts}, existing approaches tend to yield results and explanations solely on \textit{superficial temporal similarity or local pattern matching}, hardly generating reliable and evidence-based reasoning (as shown in Figure~\ref{fig_intro}). 

The underlying reason of the above problem is not the insufficient accessed data or model capacity, but \textbf{the lack of explicit rationale priors empowering in-context learning for time series reasoning}.
Thus, we propose \textit{rationale-grounded in-context learning for time series reasoning}. 
Each rationale consists of structured reasoning paths, connecting the observable cross-variable coordination with specific downstream implications, which work as reasoning guides for in-context learning rather than post-hoc explanations of given outcomes.
By grounding in-context reasoning on these rationales, MLLMs can deduce why particular temporal contexts lead to some outcomes, making MLLMs less prone to hallucinated or unjustified conclusions.

Given this insight, we introduce \texttt{RationaleTS}, a novel method that enhances time series reasoning ability of MLLMs with rationale-grounded in-context learning (Figure~\ref{fig_intro}). 
We first induce ground truth-conditioned rationales to build reasoning paths between cross-variable observations and implications on specific outcomes.
We then design a hybrid retrieval mechanism to retrieve guiding rationales for a given sample, balancing both temporal patterns and semantic contexts.
Finally, we complement the in-context inference of MLLMs with these rationale priors for evidence-grounded outcome predictions and interpretations.
Note that our method differs fundamentally from existing Retrieval-Augmented Generation (RAG)~\cite{lewis2020retrieval,jiang2023active,zheng2025retrieval} or exemplar-based In-Context-Learning (ICL)~\cite{wei2022chain} prompting methods. 
\texttt{RationaleTS} retrieves guiding rationales, instead of referenced samples, labels, or factual knowledge, which provides prior reasoning paths for in-domain tasks.
Our contributions are summarized as follows:
\begin{itemize}[leftmargin=*]
    \item We identify the key limitation of existing MLLMs for time series reasoning as the absence of rationale priors that connect temporal observations to their downstream outcomes.
    \item We introduce the rationale-grounded in-context learning for time series reasoning, which treats rationales as guiding reasoning units rather than post-hoc explanations.
    \item We propose \texttt{RationaleTS}, which induces label-consistent rationales and retrieves temporal-and-semantic similar rationale priors for in-context reasoning of MLLMs on new samples.
    \item The extensive experiments across three real-world time series reasoning datasets demonstrate that grounding reasoning on rationales promises improved effectiveness and efficiency.
\end{itemize}

\section{Related Works}

\subsection{LLMs and MLLMs for Time Series}
Given the reasoning and interpretability of LLMs, existing works have attempted to bring such ability to the time series community~\cite{jin2023time,zhou2023one,gruver2023large,liu2024time,xie2024chatts,wang2025chattime,zhong2025time}. Most of these works treat time series data as numerical sequences and try to tackle the problem of modality alignment by tokenization, reprogramming and prompt engineering~\cite{rasul2023lag,cheng2025instructime,ni2025harnessing}. However, LLMs struggle with capturing series-level contexts due to limited horizon windows.
MLLMs provide a promising way to visualize the numerical data, which can align the time steps of different variables~\cite{openai2024gpt4technicalreport,comanici2025gemini25pushingfrontier,liu2025picture}. Recent works on MLLMs mainly concentrate on chart comprehension and value perceptions, instead of complicated temporal reasoning~\cite{zhou2024can,zhang2025insight}.

\begin{figure*}[!t]
    \centering
    \includegraphics[width=0.9\linewidth]{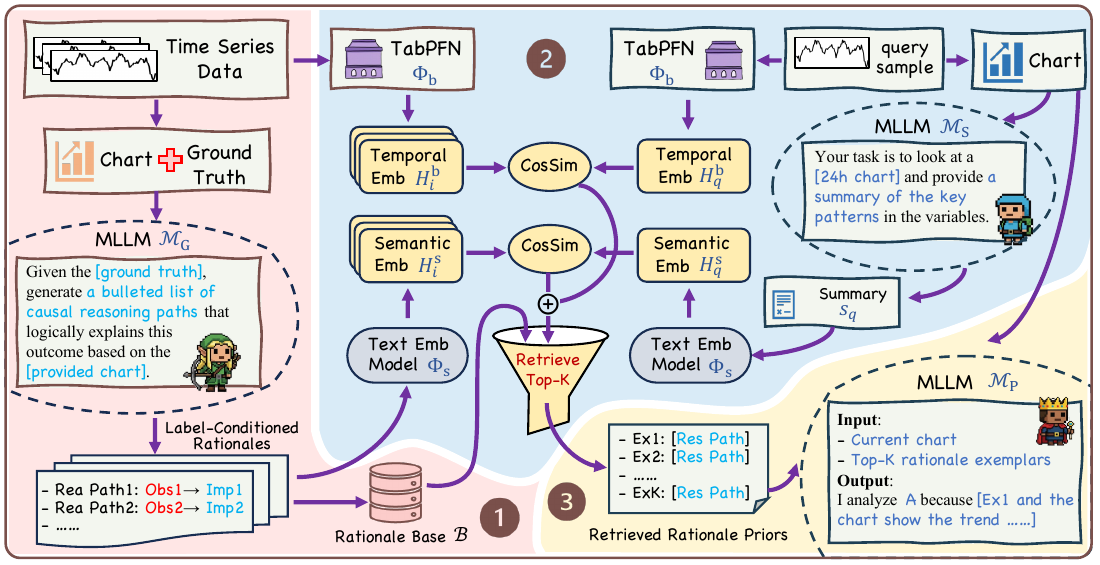}
    \caption{The workflow of \texttt{RationaleTS}, which includes \ding{202} Abductive Rationale Generation (\S \ref{sec_gen}), \ding{203} Hybrid Retrieval (\S \ref{sec_hybrid}), and \ding{204} Rationale-Grounded In-Context Inference (\S \ref{sec_infer}). }
    \label{fig_method}
\end{figure*}

\subsection{Rationale Generation}
The existing works have suggested that explicit rationales can enhance the reasoning ability of LLMs, compared with just true answers~\cite{wei2022chain,zhang2024multimodal}. 
Prior works propose to treat the generated rationales as supervision signals for training small models~\cite{hsieh2023distilling,wang-etal-2023-scott}.
For the label-conditioned rationale generation, we can track back to \cite{camburu2018snli}, where human explanations are collected conditioned on gold labels.
While in STaR~\cite{zelikman2022star}, a bootstrapping approach is introduced, where the LLM generates rationales conditioned on the correct answer to enlarge the fine-tuning datasets for iteratively training itself. 
\cite{chen-etal-2023-zara} automatically aligns the generated rationales with the correct answers and thus constructs the self-training datasets for small language models.
These methods primarily aim to generate rationales for the downstream fine-tuning the model itself or student models~\cite{shinn2023reflexion,madaan2023self,liu2024chain}. While in our work, label-conditioned rationales include reasoning paths from the temporal variable changes to the implications. The rationales can constitute high-quality knowledge base for in-context learning rather than fine-tuning models.

\section{Methodology}

\subsection{Problem Formulation}
Let $\mathcal{D}=\{(X_i, y_i)\}_{i=1}^D$ denote the time series dataset, where each sample $X_i \in \mathbb{R}^{T\times N}$ has $T$ historical time steps and $N$ variables. 
$y_i$ may represent continuous values in future steps or the discrete future trend.
Following \cite{jiang2025timexl,lee2025timecap}, we focus on the latter one and $y_i \in \mathbb{Z}_{\ge 0}$ denotes the discrete classification of a variable's future trend (e.g., 0: ``increase''; 1: ``decrease''; 2: ``stable'').
It is more beneficial to decision-critical applications, where early warnings rely on the evaluation of future trend direction, compared with accurate but uncertain continuous predictions.
Instead of simple classification, we aim to excavate the synergistic effects of different variables from historical contexts and yield results and explanations via rationale-grounded in-context learning. 


\begin{figure*}[!t]
    \centering
    \includegraphics[width=0.8\linewidth]{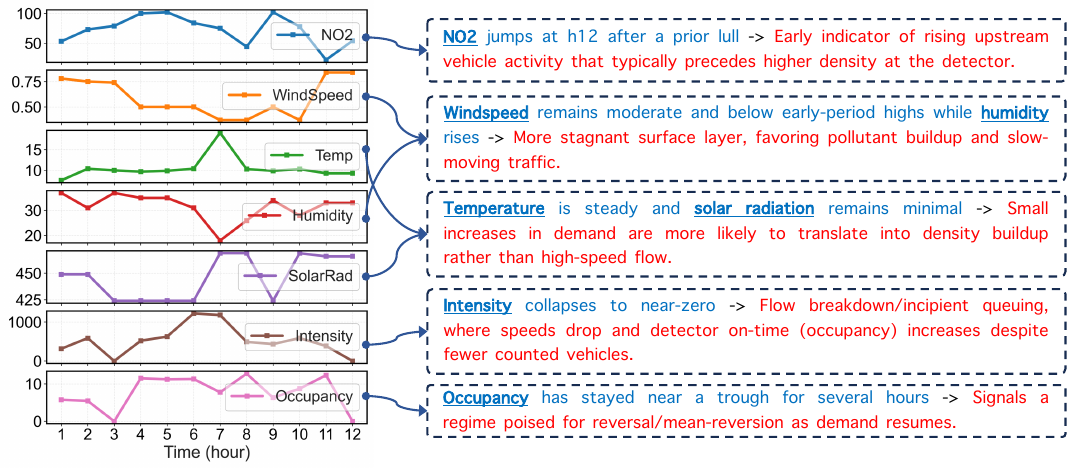}
    \caption{\textbf{Left}: Time series chart of a sample from Traffic dataset. \textbf{Right}: The generated rationales include 5 reasoning paths. \textcolor{blue}{Blue}: Observations. \textcolor{red}{Red}: Implications. Each reasoning path provides the evidence-grounded analysis on implications to the final outcome. }
    \label{fig_reason}
\end{figure*}

\subsection{\texttt{RationaleTS}}

The workflow of our proposed \texttt{RationaleTS} is shown in Figure~\ref{fig_method}. 
In the process of \textbf{abductive rationale generation}, we concatenate each time series chart and corresponding ground-truth labels to encourage the MLLM to generate hindsight reason paths, thus providing the guiding rationale base for downstream in-domain reasoning tasks.
We then propose a \textbf{hybrid retrieval} mechanism to retrieve Top-$K$ label-free rationale priors, which balances temporal patterns and semantic contexts, for the final rationale-grounded \textbf{in-context inference}.


\subsubsection{Abductive Rationale Generation}
\label{sec_gen}
The pretrained MLLMs have naive understanding of domain-specific knowledge and may result in hallucination issues, hardly learning the synergistic effects of variables and the contextual information for complicated reasoning.
Inspired by \cite{zelikman2022star}, we propose an abductive rationale generation mechanism, where the MLLM is tasked with justifying the ground-truth results by constructing evidence-grounded reasoning paths.

For each time series sample $X_i$, we firstly obtain the corresponding visual chart, denoted as $X_i^{\rm c}$. 
We concatenate $X_i^{\rm c}$ and $y_i$ to encourage the pretrained MLLM $\mathcal{M}_{\rm G}$ to generate label-conditioned rationales $r_i$, i.e., $r_i \gets \mathcal{M}_G(X_i^{\rm c}, y_i)$.
The process can be seen as a conditional text generation task. The rationale $r_i$ (supposing including $\mathcal T_i$ tokens) is generated autoregressively as:
\begin{equation}
P(r_i \mid X_i^{\rm c}, y_i) = \prod_{t=1}^{\mathcal{T}_i} P_{\mathcal{M}_{\rm G}}(r_{i,t} \mid X_i^{\rm c}, y_i, r_{i,<t}). \notag
\end{equation}
The label-conditioned rationales are organized as a bulleted list of reasoning paths following the format of ``$\mathtt{Observation \to Implication}$'', which describe how different variables synergize towards future results, as shown in Figure~\ref{fig_reason}. It is beneficial to construct a high-quality rationale database, i.e., $\mathcal{B}=\{r_i\}_{i=1}^D$ for in-context inference.
Note that we avoid revealing the true labels in rationales to encourage principled reasoning rather than simple imitation (detailed results are provided in \S \ref{sec_ablation}). 

\subsubsection{Hybrid Retrieval}
\label{sec_hybrid}
The effectiveness of in-context inference largely hinges on the retrieved exemplars (rationales in our paper)~\cite{brown2020language,alayrac2022flamingo,wang-etal-2024-learning}. 
In time series reasoning tasks, a natural way is to retrieve the rationales of time series with similar temporal patterns, which however lacks semantic contexts of observations.
For example, the decrease in the density of PM 2.5 in some cases may result from either higher wind speed or seasonality.
Besides, text summary provides coarse-grained abstraction but not quantitative details, which may aggravate hallucination issues.
Thus, we propose a hybrid retrieval approach to unify the statistical priors and semantic contexts.

\paragraph{\textbf{Data-Centric Similarity.}}
The accuracy of time series analysis depends on how to model the synergies in different variables, while the existing time series foundation models mainly leverage the channel-independent strategy~\cite{nie2022time,woo2024unified,ansari2024chronos}.
Given the effectiveness of TabPFN in time series forecasting~\cite{hoo2024the}, we leverage the frozen representation power of TabPFN as a universal time-series encoder to obtain temporal embedding.

Specifically, we reorganize raw time series $X_i$ into tabular data $X_i^{\rm b}$, with each column and row corresponding to a variable and a time step respectively. Similarly, for a $query$ \footnote{We name a $test$ sample a $query$ here~\cite{an2025thread}.} sample $X_q$, we have $X^{\rm b}_q$. Let $\Phi_{\rm b}(\cdot)$ denote the TabPFN encoder. We can obtain the temporal embeddings as:
\begin{equation}
    {H}_q^{\rm{b}} = \Phi_{\mathrm b}(X^{\mathrm b}_q), {H}_i^{\mathrm{b}} = \Phi_{\mathrm b}(X^{\mathrm b}_i).
\end{equation}
${H}_q^{\mathrm{b}}$ and $H_i^{\mathrm{b}}$ provide data priors on the coordination of temporal variables.
We then compute the data-centric similarity as:
\begin{equation}
    Sim^{\mathrm b}_i = \mathbf{cos}(H_q^{\mathrm{b}}, H_i^{\mathrm{b}}) = \frac{H_q^{\mathrm{b}} \cdot H_i^{\mathrm{b}}}{\left\lVert H_q^{\mathrm{b}} \right\rVert \left\lVert H_i^{\mathrm{b}} \right\rVert}.
\end{equation}

\paragraph{\textbf{Semantic-Centric Similarity.}}
The generated factual rationales in \S \ref{sec_gen} promise high-quality semantic contexts, including observations and potential effects. 
However, the proposed abductive rationale generation will not work for query samples without ground-truth labels. 
Therefore, 
we adopt an intermediate MLLM $\mathcal{M}_{\rm S}$ to generate the text summary of a query chart, which aligns with rationales in both modality and semantic space. 
Note that $\mathcal{M}_{\rm S}$ merely abstracts temporal changes, i.e., providing the ``\texttt{Observations}'' in reasoning paths.
Compared with complicated reasoning, this is a simple task $\mathcal{M}_{\rm S}$ can handle.

\begin{table*}[!t]
\footnotesize
\centering
\tabcolsep=0.08cm
\renewcommand\arraystretch{0.8}

\caption{Comparison of time series reasoning paradigms.}
\label{tab_comp}

\begin{tabular*}{\textwidth}{c @{\extracolsep{\fill}} cccc}
\toprule
\textbf{Method} & 
\textbf{Reasoning Unit} & 
\textbf{Reasoning Prior} & 
\textbf{In-Context Usage} & 
\textbf{Limitation} \\
\midrule
ICL 
& Exemplars
& Implicit 
& Retrieved exemplars 
& Focusing on sample similarity\\
\midrule
RAG
& Retrieved knowledge 
& Implicit
& Retrieved documents 
& Not for Time series reasoning\\
\midrule
\makecell{Rationale-\\supervised} 
& \makecell{Rationale-\\as samples} 
& Fixed 
& / 
& \makecell{Hardly connecting reasoning \\ to rationales} \\

\midrule
\texttt{RationaleTS}
& Rationales
& \textbf{Explicit}, \textbf{Transferrable}
& Retrieved rationales
& --- \\
\bottomrule
\end{tabular*}
\end{table*}

Specifically, given a query chart $X_q^{\rm c}$, the text summary is generated as $s_q \gets \mathcal{M}_{\rm S}(X_q^{\rm c})$. 
Let $\Phi_{\mathrm e} (\cdot)$ denote the text embedding model. We can obtain the semantic embeddings and evaluate the semantic-centric similarity as:
\begin{equation}
    {H}_q^{\mathrm{s}} = \Phi_{\mathrm s}(s_q), {H}_i^{\mathrm{b}} = \Phi_{\mathrm b}(r_i).
\end{equation}
\begin{equation}
    Sim^{\mathrm s}_i = \mathbf{cos}(H_q^{\mathrm{s}}, H_i^{\mathrm{s}}).
\end{equation}

\paragraph{\textbf{Hybrid Fusion.}}
 To unify the statistical priors and semantic contexts, we combine the above similarity scores with a balancing factor $\lambda$:
\begin{equation}
    Sim_i^{\mathrm{final}} = \lambda \cdot Sim_i^{\mathrm b} + (1-\lambda) \cdot Sim_i^{\mathrm s}.
\end{equation}
We construct $\mathcal{R}$ by retrieving rationales from $\mathcal{B}$ with top $K$ highest hybrid similarity scores:
\begin{equation}
\mathcal{R} = \{r_i \mid i \in \operatorname{arg\,top\text{-}K}_{i \in [1, \dots, D]} Sim_i^{\text{final}}\}.
\end{equation}

\subsubsection{In-Context Inference}
\label{sec_infer}
The retrieved explicit rationales empower the MLLM $\mathcal{M}_{\rm P}$ to perform in-context inference, by transferring the logical deduction patterns from these reasoning priors to the new query sample.
Specifically, we concatenate the query chart $X_q^{\rm c}$ and rationale set $\mathcal{R}$ as augmented input and the inference process of $\mathcal{M}_{\rm P}$ is formulated as:
\begin{equation}
    (\hat{r}_q, \hat{y}_q) = \arg\max_{r,y} P_{\mathcal{M}_{\rm P}}(r, y \mid \mathcal{R}; X_q^{\rm c}).
\end{equation}
Typically, this is decomposed into a two-step generation process:
\begin{align}
     P_{\mathcal{M}_{\rm P}}&(r, y \mid \mathcal{R}; X_q^{\rm c}) 
    \notag
    \\
    &=
\underbrace{P_{{\mathcal{M}_p}}(r \mid \mathcal{P})}_{\text{Reasoning Generation}}
\underbrace{P_{{\mathcal{M}_p}}(y \mid \mathcal{P}, r)}_{\text{Final Result}}.
\end{align}
We bootstrap $\mathcal{M}_{\rm P}$ to generate reasoning first and then results, which ensures the final inference logically consistent with the visual evidence and contextual knowledge from rationale priors.
The detailed process of \texttt{RationaleTS} is provided in Appendix~\ref{app_alg}.

\subsection{Method Analysis}
We compare different time series reasoning paradigms in Table~\ref{tab_comp}. 
In ICL~\cite{wang-etal-2024-learning} or RAG~\cite{jiang2023active,zheng2025retrieval} paradigms, the models are augmented by retrieved units, which provides similar samples but implicit reasoning priors.
While in rationale-supervised methods~\cite{shinn2023reflexion,madaan2023self,liu2024chain}, rationales work as supervised signals for fine-tuning models, where the reasoning process is not grounded on rationales.
Our proposed \texttt{RationaleTS} goes beyond the above limitations, by grounding in-context reasoning on explicit rationales which contain reasoning paths from observations to implications.

\section{Experiments}

\subsection{Experimental Settings}


\textbf{Datasets and Tasks.} 
We evaluate the performance of $\texttt{RationaleTS}$ on datasets of three domains: finance, transportation, and energy. These datasets all include multiple variables and pose complicated time series reasoning tasks. The datasets and the corresponding tasks are described as follows. More details are presented in Appendix~\ref{app_datasets}.

\textbf{Finance} includes the daily records of 9 financial indicators from January 2019 to December 2023~\cite{lee2025timecap}. Our task is to reason the S\&P 500 in the next day will \textit{increase by over 1\%, decrease by over 1\%, or remain stable} w.r.t the last day of a given 20-day period.
\textbf{Traffic} includes the hourly records of 7 weather and transportation indicators from January 2019 to June 2019~\cite{iskandaryan2022bidirectional}.
The task is to reason the occupancy of the next hour, w.r.t the last hour of a 12-hour period, will \textit{increase by 2, decrease by 2, or remain stable}.
\textbf{Power} includes 10-min records of 9 variables from meteorologic system and wind turbine SCADA in 2021~\cite{zhou2024sdwpf}. We aim to infer whether the average active power in the next 6 hours \textit{will surpass} that of the past 24 hours.

\paragraph{\textbf{Baselines.}}
We conduct the comparison experiments with three types of baselines. 
\textbf{(1) Time series reasoning methods with LLMs}: Moirai~\cite{woo2024unified}, ChatTS~\cite{xie2025chatts}, ChatTime~\cite{wang2025chattime}, 
and TimeXL~\cite{jiang2025timexl};
\textbf{(2) VL-Time}~\cite{liu2025picture} \textbf{with different base MLLMs}: GPT-4o~\cite{hurst2024gpt}, GPT-5\footnote{https://cdn.openai.com/gpt-5-system-card.pdf}, gemini-2.0-flash~\cite{team2023gemini}, gemini-2.5-flash~\cite{comanici2025gemini}, qwen-vl-max~\cite{bai2025qwen2}, and qwen3-vl-plus~\cite{yang2025qwen3}.
\textbf{(3) We evaluate the performance of a same MLLM (\texttt{GPT-4o-mini}) in the textual and visual modality}. We also augment the MLLM with CoT~\cite{wei2022chain} and In-Context Learning (ICL)~\cite{wang-etal-2024-learning}. The detailed prompts are provided in Appendix~\ref{app_prompt}.

\paragraph{\textbf{Implementations.}}
We employ \texttt{GPT-5} to generate rationales and \texttt{GPT-4o-mini} to generate text summary and perform the final prediction.
We adopt \texttt{text-embedding-3-large} as the text embedding model.
The datasets are divided with the ratio of 8:2. We construct rationale base on the 80\% samples and perform in-context inference on the 20\%.
For fair comparison, we report the performance of \texttt{RationaleTS} and the zero-shot baselines on the 20\% samples. 
We evaluate the performance of all methods with the widely-used F1 score and AUC.
The parameters of $K$ and $\lambda$ are set as 5 and 0.8.


\begin{table}[!tbp]
\small
\centering
\tabcolsep=0.1cm
\renewcommand\arraystretch{1.2}
\caption{Time series reasoning performance comparison. \textbf{\textcolor{red}{Bold}}: the best. \textcolor{blue}{\underline{Underline}}: the second best.}
\label{tab_main}
\rowcolors{3}{white}{gray!15}

\begin{tabular}{lcccccc}
\toprule
\rowcolor{white} 
 Dataset & \multicolumn{2}{c}{Finance} & \multicolumn{2}{c}{Power} & \multicolumn{2}{c}{Traffic} \\
\rowcolor{white} 
                 Metric & F1           & AUC          & F1          & AUC         & F1           & AUC          \\ \hline
Moirai            & 36.57        & 53.44        & 50.35       & 54.23       & \textcolor{blue}{\underline{62.23}}        & \textcolor{blue}{\underline{71.81}}        \\
ChatTS            & 30.30        & 50.54        & 43.26       & 52.63       & 18.27        & 44.31        \\
ChatTime          & 6.74         & 50.00        & 28.33       & 50.49       & 10.79        & 50.00        \\
TimeXL            & 19.61        & 50.73        & 67.39       & 67.50       & 20.07        & 41.36        \\ \hline
GPT-4o            & 55.65        & 66.73        & 68.02       & 67.83       & 37.93        & 53.45        \\
GPT-5             & \textcolor{blue}{\underline{66.53}}        & \textcolor{blue}{\underline{74.90}}        & 69.50       & 69.32       & 61.38        & 71.03        \\
gemini-2.0-flash  & 22.58        & 41.94        & 58.50       & 61.67       & 47.59        & 60.69        \\
gemini-2.5-flash  & 28.40        & 50.00        & 41.00       & 50.00       & 48.97        & 61.72        \\
qwen-vl-max       & 44.35        & 58.27        & \textcolor{blue}{\underline{69.61}}       & 69.41       & 9.84         & 33.11        \\
qwen3-vl-plus     & 43.55        & 57.66        & 70.60       & \textcolor{blue}{\underline{70.15}}       & 13.79        & 35.34        \\ \hline
textual           & 24.60        & 43.45        & 52.50       & 55.84       & 32.41        & 49.31        \\
textual+CoT       & 30.59        & 50.27        & 66.50       & 64.17       & 37.93        & 53.45        \\
textual+ICL       & 29.84        & 47.38        & 64.00       & 61.12       & 36.55        & 52.41        \\
visual            & 45.16        & 58.87        & 63.00       & 55.62       & 40.69        & 55.52        \\
visual+CoT        & 51.61        & 63.71        & 67.00       & 61.24       & 42.07        & 56.55        \\
visual+ICL        & 62.50        & 71.88        & 65.50       & 62.02       & 44.14        & 58.10        \\ \hline
\texttt{RationaleTS}             & \textbf{\textcolor{red}{69.76}} & \textbf{\textcolor{red}{77.32}} & \textbf{\textcolor{red}{71.50}} & \textbf{\textcolor{red}{72.87}} & \textbf{\textcolor{red}{66.21}} & \textbf{\textcolor{red}{74.66}} \\ \toprule
\end{tabular}
\end{table}

\subsection{Main Results}
Table~\ref{tab_main} shows the numerical results of the proposed \texttt{RationaleTS} and different baselines. 
The baselines in Type (1) perform poorly in time series reasoning, where the tokenization of numerical data can hardly reserve the intrinsic temporal patterns, thus affecting the learning of coordination of different variables.
In contrast, MLLMs have general better understanding of multi-variable time series. 
By comparing models of the same series, higher accuracy is obtained with higher edition, which indicates the scaling laws retain in MLLMs for time series reasoning~\cite{kaplan2020scaling}.
In baselines of Type (3), we respectively input the numerical data and visual plots to the same MLLM, i.e., \texttt{GPT-4o-mini}. 
The key motivation is that the visualization can augment the time series reasoning capability of MLLM, with the F1 score increasing by 13.11\% on average. 
Moreover, on two modalities, both CoT and ICL can improve the reasoning performance.
Our proposed \texttt{RationaleTS} outperforms on all datasets, indicating the the effectiveness of in-context inference, grounded on the high-quality rationale base and hybrid retrieval.


\begin{table}[!ht]
\small
\centering
\tabcolsep=0.3cm
\renewcommand\arraystretch{1.1}
\caption{Ablation results on Finance and Power datasets. \textcolor{bestRed}{\textbf{Bold}}: the best. \textcolor{blue}{\underline{Underline}}: the second best.}
\label{tab_ablation}
\begin{tabular}{lccccc}
\toprule
Datasets & \multicolumn{2}{c}{Finance} & \multicolumn{2}{c}{Power} \\ 
Variants  & F1 & AUC & F1 & AUC \\ 
\midrule

\rowcolor{groupC}
$\text{A.1}_\text{ w/ chart}$ & 64.11 & 73.08 & 62.50 & 63.38 \\
\rowcolor{groupC}
$\text{A.2}_\text{ w/ label}$& 64.92 & 73.69 & \textcolor{blue}{\underline{68.00}} & \textcolor{blue}{\underline{67.49}} \\
\rowcolor{groupC}
$\text{A.3}_\text{ w/ both}$& 62.50 & 71.88 & 64.50 & 65.08 \\

\rowcolor{groupD}
$\text{B.1}_\text{ w/o data}$ & 64.11 & 70.38 & 52.50 & 58.82 \\
\rowcolor{groupD}
$\text{B.2}_\text{ w/o semantic}$ & \textcolor{blue}{\underline{66.13}} & \textcolor{blue}{\underline{74.60}} & 66.50 & 64.73 \\
\rowcolor{groupD}
$\text{B.3}_\text{ random}$& 57.08 & 60.92 & 63.50 & 59.21 \\

\texttt{RationaleTS} &  \textcolor{red}{\textbf{69.76}} & \textcolor{red}{\textbf{77.32}} & \textcolor{red}{\textbf{71.50}} & \textcolor{red}{\textbf{72.87}} \\
\bottomrule
\end{tabular}
\end{table}

\begin{figure*}[!t]
    \centering
    \includegraphics[width=0.92\linewidth]{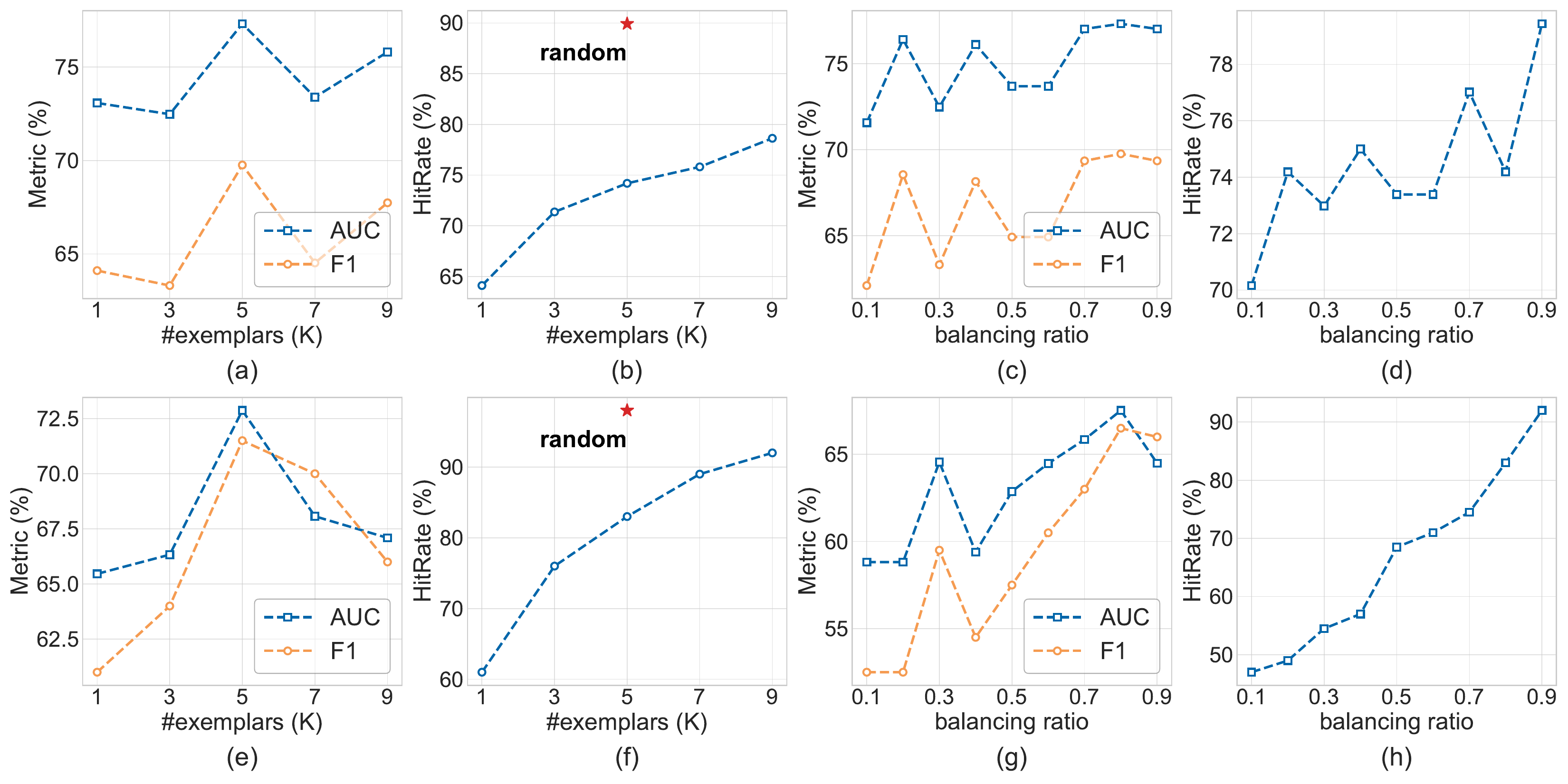}
    \caption{Hyperparamter analysis of $K$ and $\lambda$ on Finance ((a)-(d)) and Power ((e)-(h)) datasets. }
    \label{fig_params}
\end{figure*}

\subsection{Analysis}

\subsubsection{Ablation Study}
\label{sec_ablation}

The ablation results are reported in Table~\ref{tab_ablation}. 
In \textbf{A.1}-\textbf{A.3}, we integrate the visual charts, ground-truth labels, or both into rationales for in-context inference.
Neither of the two can improve the reasoning performance.
The disclosure of labels may induce the MLLM to directly copy the results, instead of decision after reasoning. 
On the other hand, the visual charts make the MLLM trapped into local pattern matching, which may provide the opposite evidence against true outcomes.

In \textbf{B.1} and \textbf{B.2}, we ablate the data-centric and semantic-centric similarity respectively.
A key observation is that ablating either would decrease the effectiveness of retrieval and ultimately impact the reasoning performance.
In \textbf{B.3}, we randomly select 5 most similar rationales, which decreases the F1 score by 12.68\% and 8\% respectively on two datasets.
The outperformance of \texttt{RationaleTS} w.r.t \textbf{B.1-B.3} demonstrates the proposed hybrid retrieval mechanism can unify statistical priors and semantic contexts and retrieve high-quality rationales for in-context inference.

\subsubsection{Sensitivity Investigation}
We conduct the sensitivity investigation of the number of rationales $K$ and balancing ratio $\lambda$ on Finance and Power datasets, as shown in Figure~\ref{fig_params}.
Besides AUC and F1 score, we adopt the $HitRate$ metric to evaluate the retrieval accuracy, which is computed as:
\begin{equation}
    HitRate = \frac{\sum_q \mathbbm{1}({\exists y_i = y_q, \forall r_i \in \mathcal{R}})}{D_q}.
\end{equation}
$\mathbbm{1}({\exists y_i = y_q, \forall r_i \in \mathcal{R}})$ denotes the indicator function, which represents whether at least one of the retrieved rationales has the same label with the query. $D_q$ denotes the number of query samples.

As shown in Figure~\ref{fig_params} (a) and (e), more rationales do not guarantee improved performance. Less rationales may not provide enough referenced knowledge to benefit reasoning, while more rationales mean more noise knowledge misleading the reasoning process. The performance is optimal when $K=5$.
As shown in Figure~\ref{fig_params} (b) and (f), the HitRate increases with more rationales. 
The HitRate is much higher in the setting of random selection. 
However, as shown in Table~\ref{tab_ablation}, \textbf{B.3} underperforms \texttt{RationaleTS} a lot, which indicates that higher HitRate may not correspond to better performance.

Figure~\ref{fig_params} (c), (d), (g), and (h) show that the three metrics have a general increasing trend with the balancing ratio $\lambda$, which indicates that data-centric similarity may play a more significant role.
\texttt{RationaleTS} achieves the best on both datasets in the setting of $\lambda=0.8$, which does not correspond to the highest HitRate, as shown in Figure~\ref{fig_params} (d) and (h).
Thus in hybrid retrieval, the semantic-centric similarity, on the other hand, can compensate for the loss of semantic contexts.

\begin{figure}[!ht]
    \centering
    \includegraphics[width=0.85\linewidth]{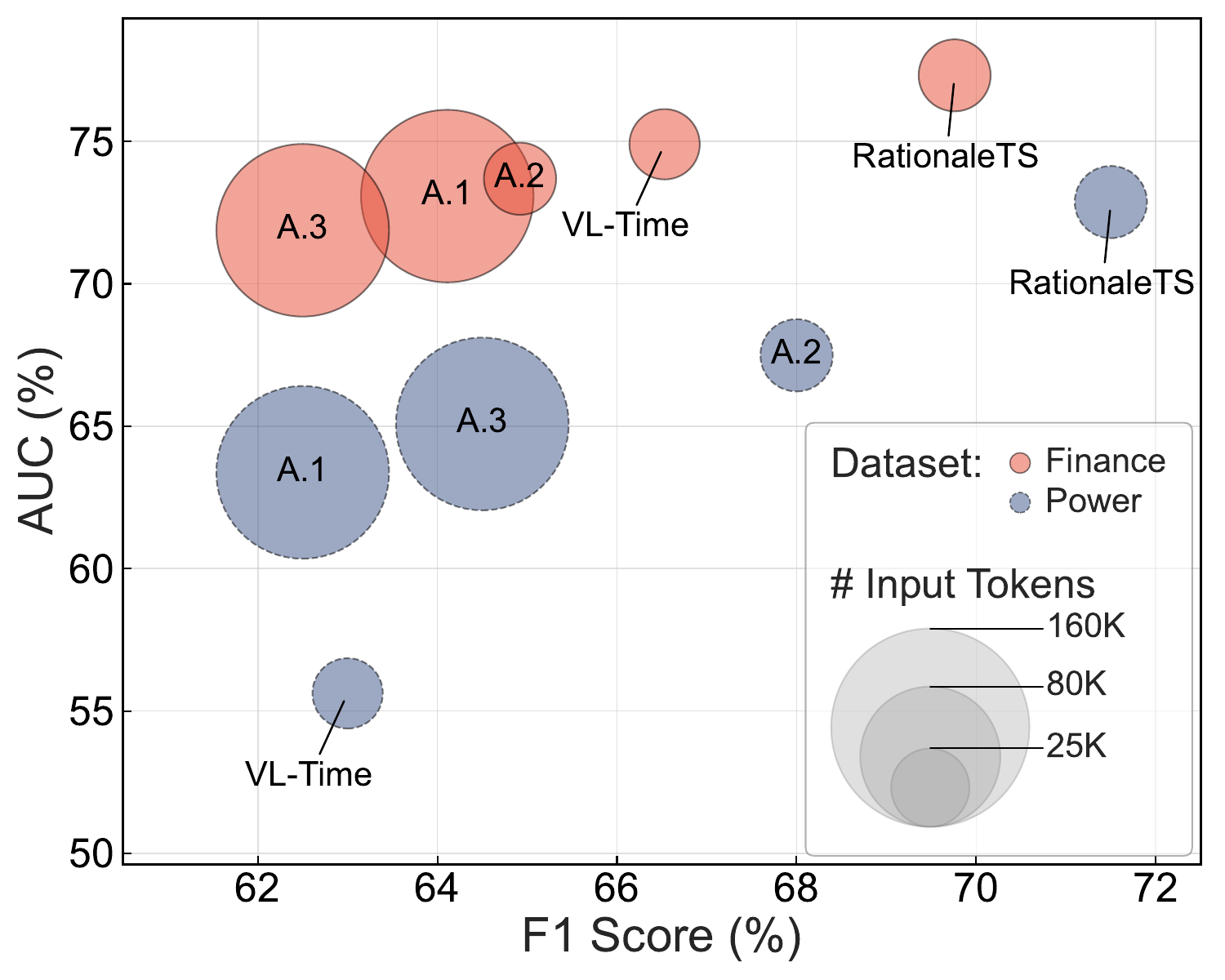}
    \caption{Efficiency analysis in terms of AUC, F1 score, and \# Input Tokens on Finance and Power datasets.}
    \label{fig_effi}
\end{figure}

\begin{figure*}[!t]
    \centering
    \includegraphics[width=\linewidth]{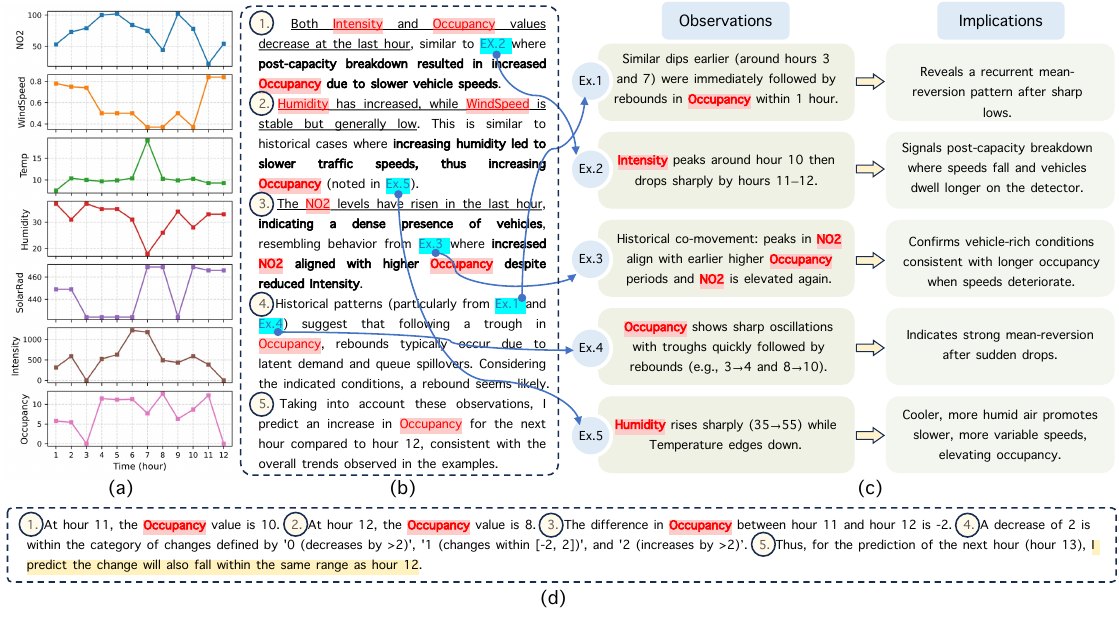}
    \caption{Case study on Traffic dataset. (a) Time series plot with 7 variables. (b) Time series reasoning of $\mathcal{M}_{\rm P}$. \underline{Underline}: Observation form the chart. \textbf{Bold}: Potential conclusion after referencing rationales. (c) The guiding reasoning paths in retrieved rationales. (d) Reasoning process of VL-Time.}
    \label{fig_case}
\end{figure*}

\subsubsection{Efficiency Analysis}
We conduct efficiency analysis by comparing the F1 score, AUC and averaged number of input tokens of $\mathcal{M}_{\rm P}$. 
Figure~\ref{fig_effi} shows the comparison among \texttt{RationaleTS}, VL-Time and three variants (in Table~\ref{tab_ablation}).
More tokens correspond to larger bubble size.
In \textbf{A.1} and \textbf{A.3}, the visual charts are incorporated into the prompts of $\mathcal{M}_{\rm P}$, which increases the tokens by 5.7$\times$ compared with \texttt{RationaleTS}. 
However, the averaged F1 score decreases by 7.23\% on two datasets.
In \textbf{A.2}, the ground-truth labels are included, which results in an averaged 4.17\% F1 decrease.
In VL-Time, $\mathcal{M}_{\rm P}$ perform zero-shot inference of each query, which has fewer tokens but 5.86\% averaged F1 score drop on two datsests.
Hence, compared with the four methods, the proposed \texttt{RationaleTS} is effective and efficient, with good balance between token usage and performance improvement.

\subsection{Case Study}
We provide the case analysis in Figure~\ref{fig_case}.
We have a key observation from Figure~\ref{fig_case} (b) that each reasoning step follows the process of \ding{202} summarizing the observations from the chart; \ding{203} seeking to specific reasoning paths from rationales for similar reasoning patterns; \ding{204} generating the potential conclusion on the future trend of \texttt{Occupancy}.
This process benefits $\mathcal{M}_{\rm P}$ to resort to specific evidence and then produce the results, avoiding the arbitrary guess and direct imitation.

Beyond co-variables, the temporal patterns of \texttt{Occupancy} in the historical horizons are also analyzed in reasoning step 4. 
The underlying reasons for the fluctuation in \texttt{Occupancy} are also analyzed.
Moreover, in reasoning step 5, the final conclusion is generated by considering the coordination of different variables, instead of merely depending on \texttt{Intensity}, which is most correlated with the targeted variable \texttt{Occupancy}.
While in Figure~\ref{fig_case}(d), VL-Time merely focuses on the changes of \texttt{Occupancy}, instead of the coordination of different variables, and produce the results following the former time stamp.
To conclude, the reasoning MLLM $\mathcal{M}_{\rm P}$ can perform effective rationale-grounded in-context inference by generating step-by-step reasoning with the process of observation, reference, and conclusion.

\section{Conclusion and Future Work}
In this paper, we identify the key limitation of MLLMs for time series reasoning tasks, namely the absence of rationale priors that connect temporal observations to their downstream implications.
We thus introduce the rationale-grounded in-context learning and propose the \texttt{RationaleTS} method, which induces label-consistent rationales and retrieves temporal-and-semantic similar rationale priors for in-context reasoning on new query samples.
Extensive experiments demonstrate the outperformance of \texttt{RationaleTS} on three reasoning tasks.
In the future work, we will explore how to construct cross-domain rationale priors and improve the rationales' structures of reasoning paths to enable more accurate retrieval.

\section*{Limitations}
The proposed method heavily depends on the generated reasoning paths in abductive rationale generation process, which are not further evaluated. 
Following existing works, this paper explores the problems of future trend prediction. More time series reasoning tasks should be considered in the future work.

\section*{Ethical Statement}

This work employs publicly available multi-modal large language models as foundational components of the proposed framework.
These models are used without additional training on private or proprietary data.
All datasets involved in our experiments are obtained from public sources and do not contain personally identifiable information.
Potential risks of this work include the misuse of forecasting results in automated decision-making systems. Our method is designed as a decision-support tool rather than a fully autonomous system.
We adopt AI Assistants for polishing the original content, rather than for suggesting new content.



\bibliography{custom}

\appendix

\section{More Related Works}

\subsection{Augmented Language Models}
The augmented language models aim to tackle hallucination issues of LLMs or MLLMs by complementing them with external knowledge for improved reasoning ability. 
\cite{wei2022chain} proposes CoT prompting, encouraging models to generate intermediate reasoning steps.
\cite{zhang2024multimodal} introduces Multimodal-CoT to vision-language domain.
RAG typically retrieves documents or simple Question-Answer pairs rather than complicated logic flows~\cite{lewis2020retrieval,jiang2023active,salemi2024evaluating,wang2024searching}. 
Furthermore, standard retrieval struggles to align the statistical properties of time series with semantic reasoning. 
Our \texttt{RationaleTS} bridges the gaps by proposing a hybrid retrieval (integrating data priors and semantic contexts) to retrieve rationale priors, enabling in-context inference.

\section{Algorithm}
\label{app_alg}
Algorithm~\ref{ag1} present the process of \texttt{RationaleTS}, with the \textcolor{blue}{blue} parts represent the construction of rationale base based on training datasets, which includes generating rationales (Line 2) and obtaining temporal and semantic embeddings (Line 3).
In the inference phase, given a query sample, the temporal and semantic embeddings are firstly obtained (Line 5-6). We then incorporate statistical priors and semantic contexts to retrieve $K$ rationales with top similarity scores (Line 7-11), based on which the in-context inference is conducted to generate the underlying reasons and results (Line 12-13).


\begin{algorithm}[!ht]
    \caption{\texttt{RationaleTS}}
    \label{ag1}
    \KwIn{
    Dataset $\{X_i, X_i^{\rm c}, X_i^{\rm b}, y_i\}_{i=1}^D$;
    Pretrained MLLMs $\mathcal{M}_{\rm G}, \mathcal{M}_{\rm S}$, and $\mathcal{M}_{\rm P}$; $\lambda$;
    Query: $(X_q, X_q^{\rm c}, X_q^{\rm b})$
    }
    \KwOut{
    $\hat{y}_q$ and $\hat{r}_q$
    }
    \For{$i \in [1, D]$}{
        {\color{blue} $r_i \gets \mathcal{M}_{\rm G} (X_i^{\rm c}, y_i)$ }

        {\color{blue} $H_i^{\mathrm{b}} = \Phi_{\mathrm b}(X^{\mathrm b}_i)$, $H_i^{\mathrm{s}} = \Phi_{\mathrm s}(r_i)$}
    }
    // \textit{In Inference Phase}
    
    $s_q \gets \mathcal{M}_{\rm S}(X_q^{\rm c})$

    $H_q^{\mathrm{b}} = \Phi_{\mathrm b}(X^{\mathrm b}_q)$, $H_q^{\mathrm{s}} = \Phi_{\mathrm s}(s_q)$ 

    \For{$i \in [1, D]$}{
        $Sim^{\mathrm b}_i = \mathbf{cos}(H_q^{\mathrm{b}}, \textcolor{blue}{H_i^{\mathrm{b}}})$
        
        $Sim^{\mathrm s}_i = \mathbf{cos}(H_q^{\mathrm{s}}, \textcolor{blue}{H_i^{\mathrm{s}}})$ 
    
        $Sim_i^{\mathrm{final}} = \lambda \cdot Sim_i^{\mathrm b} + (1-\lambda) \cdot Sim_i^{\mathrm s}$
        
    }

    $\mathcal{R} = \{r_i \mid i \in \operatorname{arg\,top\text{-}K}_{i \in [1, \dots, D]} Sim_i^{\text{final}}\}$

    $\hat{r}_q, \hat{y}_q \gets \mathcal{M}_{\rm P} (\mathcal{R}, X_q^{\rm c})$
    
    \Return $\hat{r}_q$ and $\hat{y}_q$ 
\end{algorithm}

\section{Details of Datasets}
\label{app_datasets}

We evaluate on three benchmarks covering the domains of finance, transportation, and energy. The details of the datasets are presented in Table~\ref{tab_datasets}. We provide the details in the following parts. 

\begin{table*}[!t]
    \small
    \centering
    \tabcolsep=0.2cm
    \renewcommand\arraystretch{1.2}
    \caption{Details of three benchmarking datasets.}
    \label{tab_datasets}
    \begin{tabular}{ccccccc}
        \toprule
        Dataset & Frequency & \#Variables & \#Time Stamps & Duration & \#Samples & Label Distribution \\
        \midrule
        Finance & 1-D & 9 & 1258 & 2019/1--2023/12 & 1238 & 13.78\% / 17.04\% / 69.18\% \\
        Traffic & 1-H & 7 & 4344 & 2019/1--2019/6 & 722 & 14.95\% / 52.22\% / 32.83\% \\
        Power   & 10-Min & 9 & 49760 & 2021/1--2021/12 & 997 & 42.05\% / 57.95\% \\
        \bottomrule
    \end{tabular}
\end{table*}

\begin{figure}[!ht]
    \centering
    \includegraphics[width=\linewidth]{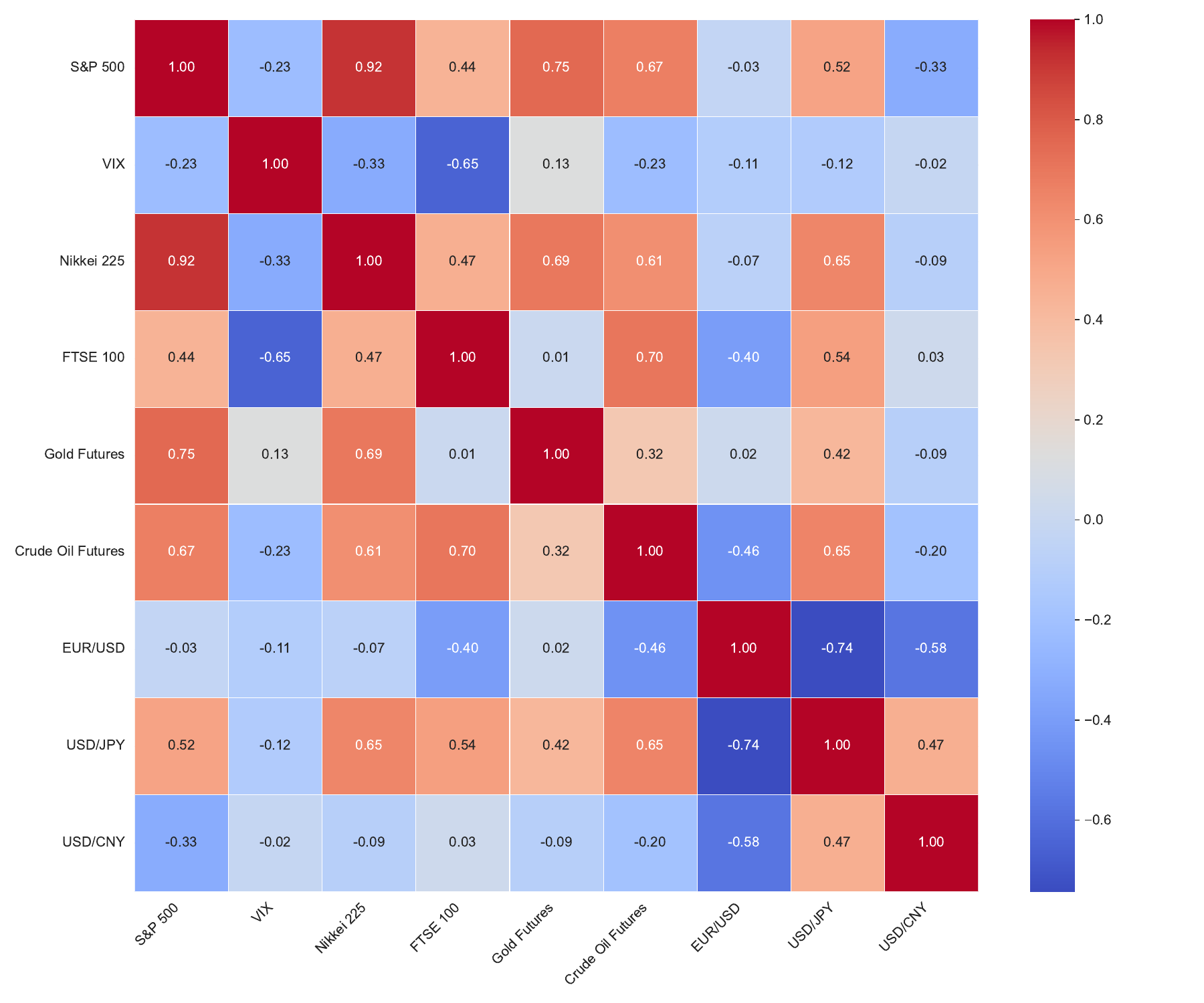}
    \caption{Correlation Matrix of 9 variables in Finance dataset.}
    \label{fig_finance}
\end{figure}

\textbf{Finance}~\cite{lee2025timecap}: 
This dataset includes 9 indicators: \textbf{\texttt{S\&P 500}, \texttt{VIX}, \texttt{Nikkei 225}, \texttt{FTSE 100}, \texttt{Gold Futures}, \texttt{Crude Oil Futures}, \texttt{EUR/USD}, \texttt{USD/JPY}, and \texttt{USD/CNY}}. As shown in Figure~\ref{fig_finance}, \texttt{S\&P 500} correlates with the other variables, especially \texttt{Nikkei 225}, \texttt{Gold Futures}, and \texttt{Crude Oil Futures}.S
Therefore, we analyze the future trend of \texttt{S\&P 500} based on the historical contexts of all nine variables. Specifically, we analyze whether the indicator of \texttt{S\&P 500} in the next one day will \textit{decrease by over 1 \% (labeled as 0), remain stable (labeled as 1), or increase by over 1 \% (labeled as 2)} w.r.t the last day of a given 20-day period.

\begin{figure}[!ht]
    \centering
    \includegraphics[width=\linewidth]{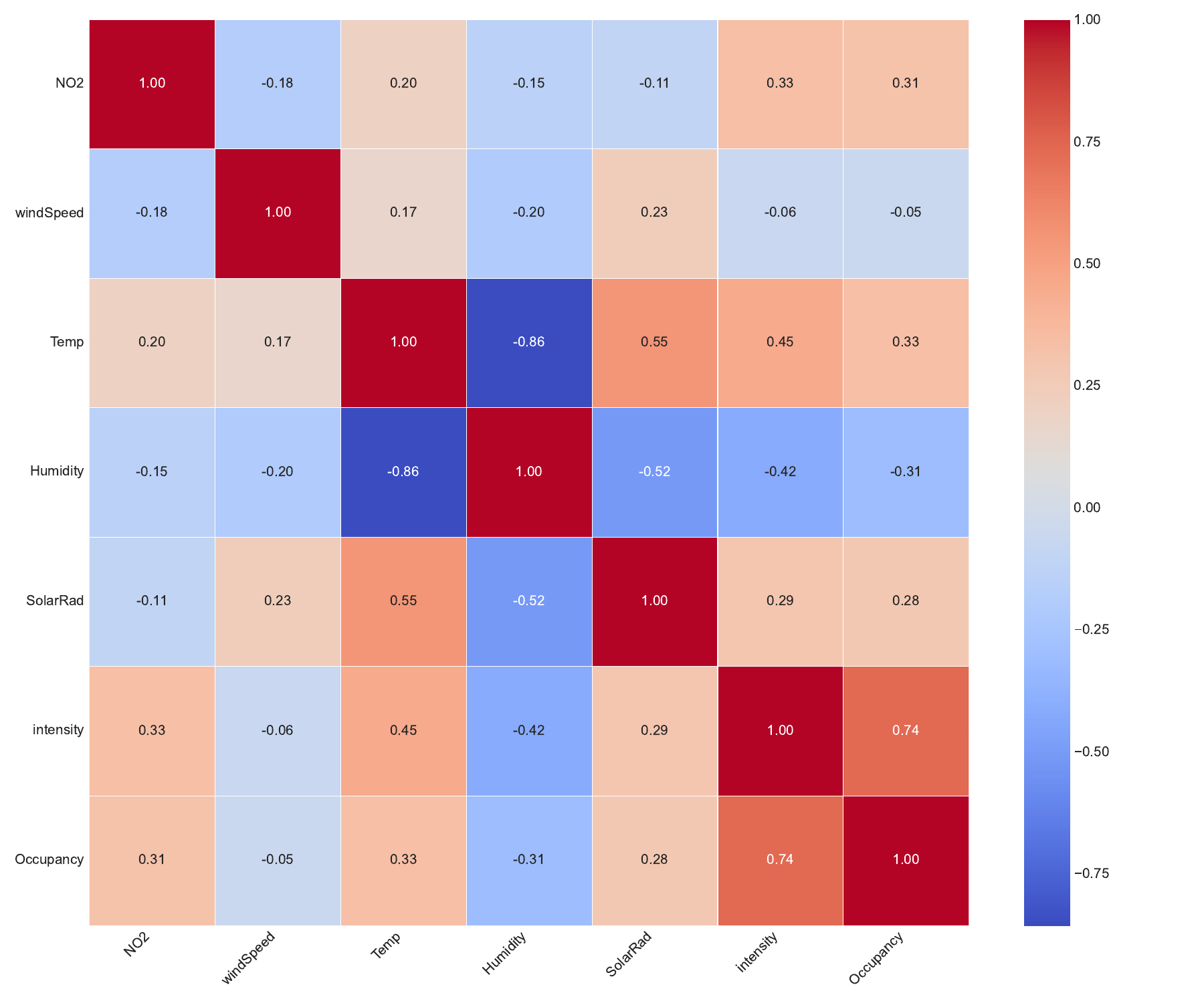}
    \caption{Correlation Matrix of 7 variables in Traffic dataset.}
    \label{fig_traffic}
\end{figure}

\textbf{Traffic}~\cite{iskandaryan2022bidirectional}: This dataset includes 5 weather indicators: \textbf{\texttt{NO2}, \texttt{WindSpeed}, \texttt{Temperature}, \texttt{Humidity}, and \texttt{SolarRad}}; and 2 transportation indicators: \textbf{\texttt{Intensity} and \texttt{Occupancy}}.
The traffic \texttt{Intensity} evaluates the number of vehicles per hour, while the \texttt{Occupancy} indicates the proportion of time that road detectors are occupied by vehicles in an hour, which reflects the traffic jam level.
We illustrate the correlation matrix of these 7 variables in Figure~\ref{fig_traffic}. 
The 5 weather indicators may affect traffic needs and thus correlated with the \texttt{Intensity} and \texttt{Occupancy}.
In this paper, we analyze the \texttt{Occupancy} of the next hour, w.r.t the last hour of a 12-hour period, will \textit{decrease by 2 (labeled as 0), remain stable (labeled as 1), or increase by 2 (labeled 2)}.

\begin{figure}[!ht]
    \centering
    \includegraphics[width=\linewidth]{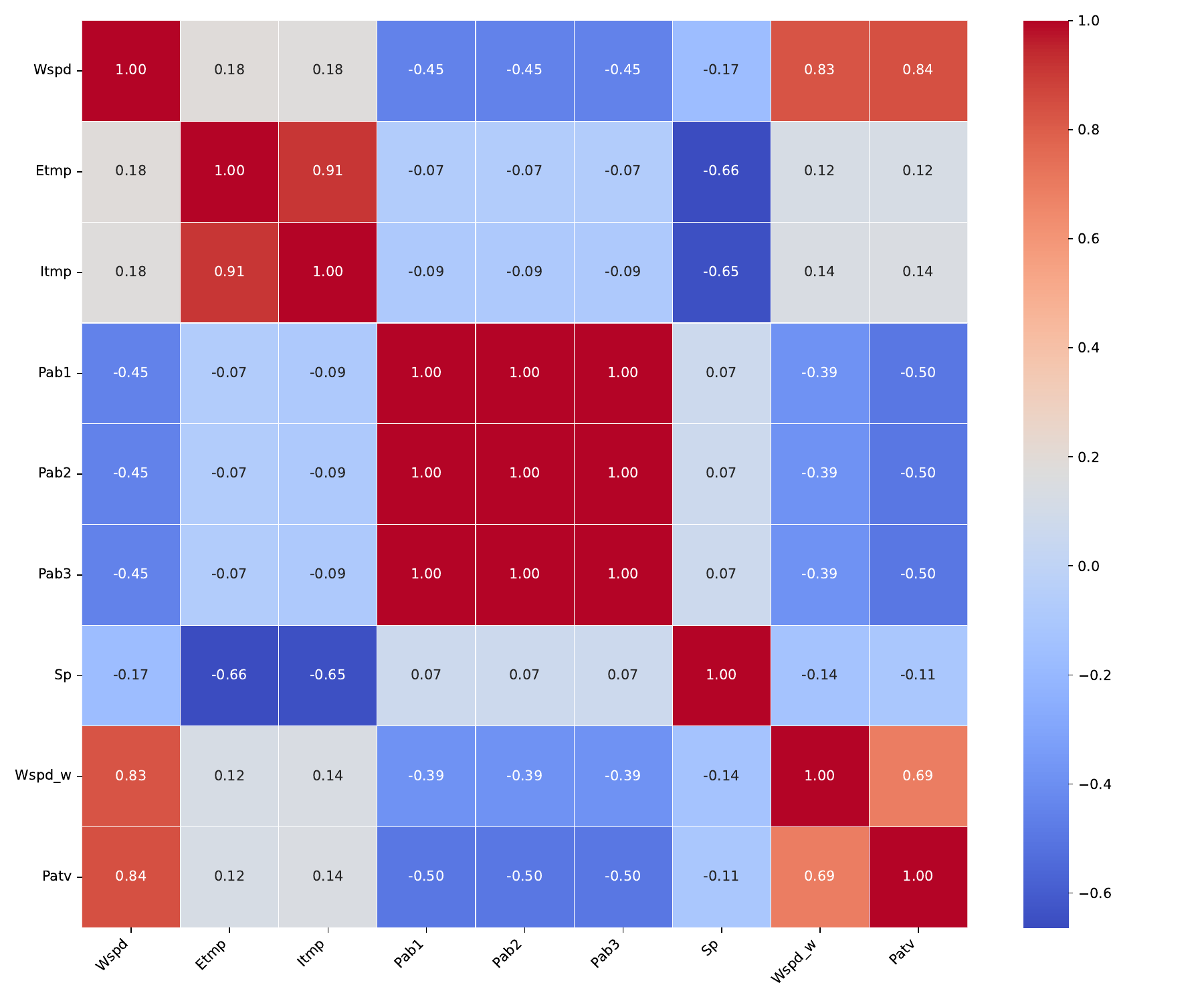}
    \caption{Correlation Matrix of 9 variables in Power dataset.}
    \label{fig_power}
\end{figure}

\textbf{Power}~\cite{zhou2024sdwpf}: This dataset includes wind speed measurements (\textbf{\texttt{Wspd}, \texttt{Wspd\_w}}), environmental and internal temperatures (\textbf{\texttt {Etmp}, \texttt{Itmp}}), blade pitch angles (\textbf{\texttt{Pab1}, \texttt{Pab2}, and \texttt{Pab3}}), rotor speed (\textbf{\texttt{Sp}}), and active power output (\textbf{\texttt{Patv}}), collectively characterizing the aerodynamic input, control actions, mechanical state, and power generation of wind turbines. Figure~\ref{fig_power} shows the correlation matrix of these 9 variables.
In this paper, we aim to analyze whether the average active power \texttt{Patv} in the next 6 hours will \textit{surpass that of the past 24 hours (labeled as 1) or not (labeled 0)}.

\section{Details of Prompts}
\label{app_prompt}
We provide the prompts of the adopted three MLLMs $\mathcal{M}_{\rm G}$, $\mathcal{M}_{\rm S}$, and $\mathcal{M}_{\rm P}$ as follows. Moreover, we also present the prompts for baselines in Type (3).

\begin{figure*}
    \begin{tcolorbox}[colback=blue!5!white, 
                  colframe=blue!75!black, 
                  colbacktitle=blue!25!white, 
                  coltitle=black,         
                  fonttitle=\bfseries,   
                  title=1. Prompt in $\mathcal{M}_{\rm G}$ for abductive rationale generation,        
                  boxrule=1pt,           
                  arc=6pt,               
                  left=0.5mm,                  
                  right=0.5mm,                 
                  top=0.5mm,
                  bottom=0.5mm
                  ]    

    \textbf{\underline{System Prompt}}:

    You are a senior [\texttt{specific domain}, \textit{e.g., traffic and urban}] analyst. Given the actual outcome, your task is to generate a concise, `gold-standard' causal reasoning path that logically explains this outcome based on the provided [\texttt{specific domain}, \textit{e.g., traffic}] chart. This path will be used for a retrieval system. **Do not mention the actual outcome or the final prediction in your reasoning text.**

    \textbf{\underline{User Prompt}}:
    
    The actual outcome for the next [future windows] was: **\{true\_label\_meaning\}**.
    
    Please provide the ideal reasoning path that explains this outcome based on the attached [historical windows] data chart.
    
    \textbf{Your Task}
    
    Provide a bulleted list of key causal factors. Each bullet point must follow the format: 'Observation -> Implication'. Focus on describing the *dynamics* and *patterns*.

\end{tcolorbox}
\end{figure*}

\begin{figure*}
    \begin{tcolorbox}[colback=blue!5!white, 
                  colframe=blue!75!black, 
                  colbacktitle=blue!25!white, 
                  coltitle=black,         
                  fonttitle=\bfseries,   
                  title=2. Prompt in $\mathcal{M}_{\rm S}$ for generating text summary,        
                  boxrule=1pt,           
                  arc=6pt,               
                  left=0.5mm,                  
                  right=0.5mm,                 
                  top=0.5mm,
                  bottom=0.5mm
                  ]    

    \textbf{\underline{System Prompt}}:
    
    You are a concise [\texttt{specific domain}, \textit{e.g., traffic}] data analyst. Your task is to look at a [\texttt{historical windows}, \textit{e.g., 12-hour}] [\texttt{specific domain}, \textit{traffic}] chart and provide a brief, factual summary of the most prominent patterns.
    
    \textbf{\underline{User Prompt}}:
    
    Analyze the attached [\texttt{historical windows}, \textit{e.g., 12-hour}] [\texttt{specific domain}, \textit{e.g., traffic}] data chart. Provide a one-paragraph summary describing the key trends you observe in variables. Be factual and objective. 

\end{tcolorbox}
\end{figure*}

\begin{figure*}
    \begin{tcolorbox}[colback=blue!5!white, 
                  colframe=blue!75!black, 
                  colbacktitle=blue!25!white, 
                  coltitle=black,         
                  fonttitle=\bfseries,   
                  title=3. Prompt in $\mathcal{M}_{\rm P}$ for in-context inference,        
                  boxrule=1pt,           
                  arc=6pt,               
                  left=0.5mm,                  
                  right=0.5mm,                 
                  top=0.5mm,
                  bottom=0.5mm
                  ]    

    \textbf{\underline{System Prompt}}:
    
    You are a world-class [\texttt{specific domain}, \textit{e.g., wind power generation}] expert. 
    
    You will be given a new [\texttt{historical windows}, \textit{e.g., 24-hour}] data chart and several relevant historical reasoning paths. 
    
    Your task is to first study the historical examples, then analyze the new chart, and finally analyze the [\texttt{targeted variable}, \textit{e.g., power output}] trend for the next [\texttt{future windows}, \textit{e.g., 6 hours}].
    
    \textbf{\underline{User Prompt}}:
        
    Here are some relevant historical reasoning paths:
    
    \{\texttt{examples}\}
    
    \textbf{Your Task}
    
    Now, analyze the **\texttt{new attached chart}**. Based on your analysis of this new chart AND the patterns learned from the historical examples, predict whether the [\texttt{targeted variable}, \textit{e.g., average active power (`Patv')}] in the next [\texttt{future windows}, \textit{6 hours}] will [\texttt{specific reasoning task}, \textit{be higher than the average of the past 24 hours}]. Categorize your prediction as [\texttt{discrete labels and meanings}, \textit{e.g., 0 (decrease by more than 1\%), 1 (remain neutral (i.e., between -1\% and 1\%)), or 2 (increase by more than 1\%)}].
    
    Provide your answer in a valid JSON format with `reasoning' and `prediction' keys. Your `reasoning' should be a step-by-step analysis that explicitly references both the new chart's data and the logic from the provided examples.

\end{tcolorbox}
\end{figure*}

\begin{figure*}
    \begin{tcolorbox}[colback=blue!5!white, 
                  colframe=blue!75!black, 
                  colbacktitle=blue!25!white, 
                  coltitle=black,         
                  fonttitle=\bfseries,   
                  title=4. Prompt for zero-shot inference in textual modality,        
                  boxrule=1pt,           
                  arc=6pt,               
                  left=0.5mm,                  
                  right=0.5mm,                 
                  top=0.5mm,
                  bottom=0.5mm
                  ]    

    \textbf{\underline{System Prompt}}:
    
    You are a world-class [\texttt{specific domain}, \textit{e.g., traffic}] expert. 
    
    You will be given [\texttt{historical windows}, \textit{e.g., 12-hour}] [\texttt{specific domain}, \textit{e.g., traffic and environment}] data. 

    Your task is to analyze the data and predict the [\texttt{targeted variable}, \textit{e.g., Occupancy}] trend for the next [\texttt{future windows}, \textit{e.g., hour}].
    
    \textbf{\underline{User Prompt}}:

    \textbf{Time-Series Data}
    
    Here is the [\texttt{historical windows}, \textit{e.g., 12-hour}] data for a specific location:
    
    [time series data]
    
    \textbf{Your Task}
    
    Analyze the provided data. Predict the change in [\texttt{targeted variable}, \textit{e.g., Occupancy}] for the next [\texttt{future windows}, \textit{e.g., hour}] compared to the last hour in the data. Categorize your prediction as [\texttt{discrete labels and meanings}, \textit{e.g., 0 (decreases by >2), 1 (changes within [-2, 2]), or 2 (increases by >2)}] .
    
    Provide your answer in a valid JSON format with `reasoning' and `prediction' keys. Your `reasoning' should be a step-by-step analysis of the data.

\end{tcolorbox}
\end{figure*}

\begin{figure*}
    \begin{tcolorbox}[colback=blue!5!white, 
                  colframe=blue!75!black, 
                  colbacktitle=blue!25!white, 
                  coltitle=black,         
                  fonttitle=\bfseries,   
                  title=5. Prompt for ICL inference in textual modality,        
                  boxrule=1pt,           
                  arc=6pt,               
                  left=0.5mm,                  
                  right=0.5mm,                 
                  top=0.5mm,
                  bottom=0.5mm
                  ]    

    \textbf{\underline{System Prompt}}:
    
    You are a world-class [\texttt{specific domain}, \textit{e.g., traffic}] expert. 

    You will be shown several examples of [\texttt{historical windows}, \textit{e.g., 12-hour}] data, each paired with its correct label indicating the [\texttt{targeted variable}, \textit{e.g., Occupancy}] change for the next [\texttt{future windows}, \textit{e.g., hour}]. Your task is to learn the patterns from these examples and then predict the change for new, unseen data.
    
    \textbf{\underline{User Prompt}}:

    Analyze the following examples. Each example consists of time-series data and its corresponding label for the [\texttt{targeted variable}, \textit{e.g., Occupancy}] change.
    
    [Example i: time series data; label meanings]

    \textbf{Your Task}
    
    Now, analyze the **new data** below. Based on the patterns you observed in the examples, predict the change in [\texttt{targeted variable}, \textit{e.g., Occupancy}] for the next [\texttt{future windows}, \textit{e.g., hour}]. 
    Categorize your prediction as [\texttt{discrete labels and meanings}, \textit{e.g., 0 (decreases by >2), 1 (changes within [-2, 2]), or 2 (increases by >2)}].
    
    \textbf{New Data}

    [time series data]
    
    Provide your answer in a valid JSON format with `reasoning' and `prediction' keys. Your reasoning should be a step-by-step analysis of the new data, drawing parallels to the provided examples where applicable.

\end{tcolorbox}
\end{figure*}

\begin{figure*}
    \begin{tcolorbox}[colback=blue!5!white, 
                  colframe=blue!75!black, 
                  colbacktitle=blue!25!white, 
                  coltitle=black,         
                  fonttitle=\bfseries,   
                  title=6. Prompt for CoT inference in textual modality,        
                  boxrule=1pt,           
                  arc=6pt,               
                  left=0.5mm,                  
                  right=0.5mm,                 
                  top=0.5mm,
                  bottom=0.5mm,
                  ]    

    \textbf{\underline{System Prompt}}:
    
    You are a world-class [\texttt{specific domain}, \textit{e.g., traffic}] expert. 

    You will be given [\texttt{historical windows}, \textit{e.g., 12-hour}] [\texttt{specific domain}, \textit{e.g., traffic}] data.
    Your task is to analyze the data and predict the [\texttt{targeted variable}, \textit{e.g., Occupancy}] for the next [\texttt{future windows}, \textit{e.g., hour}].
    
    \textbf{\underline{User Prompt}}:

    \textbf{Time-Series Data}
    
    Here is the [\texttt{historical windows}, \textit{e.g., 12-hour}] data for a specific location:
    
    [time series data]
    
    \textbf{Your Task}
    
    Analyze the provided data. Predict the change in [\texttt{targeted variable}, \textit{e.g., Occupancy}] for the next [\texttt{future windows}, \textit{e.g., hour}] compared to the last hour in the data. Categorize your prediction as [\texttt{discrete labels and meanings}, \textit{e.g., 0 (decreases by >2), 1 (changes within [-2, 2]), or 2 (increases by >2)}].
    
    Please provide the ideal reasoning path that explains your prediction based on the provided data, following the format: `Observation -> Implication`.
    
    Provide your answer in a valid JSON format with `reasoning' and `prediction' keys. Your `reasoning' should be a step-by-step analysis of the data.

\end{tcolorbox}
\end{figure*}

\begin{figure*}
    \begin{tcolorbox}[colback=blue!5!white, 
                  colframe=blue!75!black, 
                  colbacktitle=blue!25!white, 
                  coltitle=black,         
                  fonttitle=\bfseries,   
                  title=7. Prompt for zero-shot inference in visual modality,        
                  boxrule=1pt,           
                  arc=6pt,               
                  left=0.5mm,                  
                  right=0.5mm,                 
                  top=0.5mm,
                  bottom=0.5mm
                  ]    

    \textbf{\underline{System Prompt}}:
    
    You are a world-class [\texttt{specific domain}, \textit{e.g., traffic}] expert. 
    
    You will be given [\texttt{historical windows}, \textit{e.g., 12-hour}] [\texttt{specific domain}, \textit{e.g., traffic and environment}] data chart. 

    Your task is to analyze the chart and predict the [\texttt{targeted variable}, \textit{e.g., Occupancy}] trend for the next [\texttt{future windows}, \textit{e.g., hour}].
    
    \textbf{\underline{User Prompt}}:
    
    \textbf{Your Task}
    
    Analyze the **Attached Chart**. Predict the change in [\texttt{targeted variable}, \textit{e.g., Occupancy}] for the next [\texttt{future windows}, \textit{e.g., hour}] compared to the last hour in the data. Categorize your prediction as [\texttt{discrete labels and meanings}, \textit{e.g., 0 (decreases by >2), 1 (changes within [-2, 2]), or 2 (increases by >2)}] .
    
    Provide your answer in a valid JSON format with `reasoning' and `prediction' keys. Your `reasoning' should be a step-by-step analysis of the chart.

\end{tcolorbox}
\end{figure*}

\begin{figure*}
    \begin{tcolorbox}[colback=blue!5!white, 
                  colframe=blue!75!black, 
                  colbacktitle=blue!25!white, 
                  coltitle=black,         
                  fonttitle=\bfseries,   
                  title=8. Prompt for ICL inference in visual modality,        
                  boxrule=1pt,           
                  arc=6pt,               
                  left=0.5mm,                  
                  right=0.5mm,                 
                  top=0.5mm,
                  bottom=0.5mm
                  ]    

    \textbf{\underline{System Prompt}}:
    
    You are a world-class [\texttt{specific domain}, \textit{e.g., traffic}] expert. 

    You will be shown several examples of [\texttt{historical windows}, \textit{e.g., 12-hour}] data chart, each paired with its correct label indicating the [\texttt{targeted variable}, \textit{e.g., Occupancy}] change for the next [\texttt{future windows}, \textit{e.g., hour}]. Your task is to learn the patterns from these examples and then predict the change for new, unseen chart.
    
    \textbf{\underline{User Prompt}}:

    Analyze the following examples. Each example consists of a chart and its corresponding label for the [\texttt{targeted variable}, \textit{e.g., Occupancy}] change.
    
    [Example i: time series chart; label meanings]

    \textbf{Your Task}
    
    Now, analyze the **Attached Chart**. Based on the patterns you observed in the examples, predict the change in [\texttt{targeted variable}, \textit{e.g., Occupancy}] for the next [\texttt{future windows}, \textit{e.g., hour}]. 
    Categorize your prediction as [\texttt{discrete labels and meanings}, \textit{e.g., 0 (decreases by >2), 1 (changes within [-2, 2]), or 2 (increases by >2)}].
    
    Provide your answer in a valid JSON format with `reasoning' and `prediction' keys. Your reasoning should be a step-by-step analysis of the new chart, drawing parallels to the provided examples where applicable.

\end{tcolorbox}
\end{figure*}

\begin{figure*}
    \begin{tcolorbox}[colback=blue!5!white, 
                  colframe=blue!75!black, 
                  colbacktitle=blue!25!white, 
                  coltitle=black,         
                  fonttitle=\bfseries,   
                  title=9. Prompt for CoT inference in visual modality,        
                  boxrule=1pt,           
                  arc=6pt,               
                  left=0.5mm,                  
                  right=0.5mm,                 
                  top=0.5mm,
                  bottom=0.5mm
                  ]    

    \textbf{\underline{System Prompt}}:
    
    You are a world-class [\texttt{specific domain}, \textit{e.g., traffic}] expert. 

    You will be given [\texttt{historical windows}, \textit{e.g., 12-hour}] [\texttt{specific domain}, \textit{e.g., traffic}] data chart.
    Your task is to analyze the chart and predict the [\texttt{targeted variable}, \textit{e.g., Occupancy}] for the next [\texttt{future windows}, \textit{e.g., hour}].
    
    \textbf{\underline{User Prompt}}:
    
    \textbf{Your Task}
    
    Analyze the **Attached Chart**. Predict the change in [\texttt{targeted variable}, \textit{e.g., Occupancy}] for the next [\texttt{future windows}, \textit{e.g., hour}] compared to the last hour in the chart. Categorize your prediction as [\texttt{discrete labels and meanings}, \textit{e.g., 0 (decreases by >2), 1 (changes within [-2, 2]), or 2 (increases by >2)}].
    
    Please provide the ideal reasoning path that explains your prediction based on the attached chart, following the format: `Observation -> Implication'.
    
    Provide your answer in a valid JSON format with `reasoning' and `prediction' keys. Your `reasoning' should be a step-by-step analysis of the chart.

\end{tcolorbox}
\end{figure*}

\end{document}